\documentclass[preprints,article,accept,moreauthors,pdftex]{mdpi}

\usepackage[misc]{ifsym}
\usepackage{amsfonts}
\usepackage{authblk}
\usepackage{float}
\usepackage{subcaption}
\hypersetup{breaklinks=true}
\usepackage{natbib}

\Title{Anomaly Detection and Automatic Labeling for Solar Cell Quality Inspection Based on Generative Adversarial Network}

\Author{Julen Balzategui *, Luka Eciolaza and Daniel Maestro-Watson}

\AuthorNames{Julen Balzategui, Luka Eciolaza and Daniel Maestro-Watson}

\AuthorCitation{Balzategui, J.; Eciolaza, L.; Maestro-Watson, D.}

\address{
\quad Electronics and Computer Science Department, Mondragon Unibertsitatea, 20500 Arrasate, Spain; jbalzategui@mondragon.edu (J.B.); leciolaza@mondragon.edu (L.E.); dmaestro@mondragon.edu (D.M.)\\
}

\corres{Correspondence: jbalzategui@mondragon.edu;}

\abstract{Quality inspection applications in industry are required to move towards a zero-defect manufacturing scenario, with non-destructive inspection and traceability of 100 \% of produced parts. Developing robust fault detection and classification models from the start-up of the lines is challenging due to the difficulty in getting enough representative samples of the faulty patterns and the need to manually label them. This work presents a methodology to develop a robust inspection system, targeting these peculiarities, in the context of solar cell manufacturing. The methodology is divided into two phases: In the first phase, an anomaly detection model based on a Generative Adversarial Network (GAN) is employed. This model enables the detection and localization of anomalous patterns within the solar cells from the beginning, using only non-defective samples for training and without any manual labeling involved. 
In a second stage, as defective samples arise, the detected anomalies will be used as automatically generated annotations for the supervised training of a Fully Convolutional Network that is capable of detecting multiple types of faults. The experimental results using 1873 EL images of monocrystalline cells show that (a) the anomaly detection scheme can be used to start detecting features with very little available data, (b) the anomaly detection may serve as automatic labeling in order to train a supervised model, and (c) segmentation and classification results of supervised models trained with automatic labels are comparable to the ones obtained from the models trained with manual labels.}

\keyword{Anomaly detection; Electroluminescence; Solar cells; Neural Networks}

\begin{document}
\end{paracol}

\section{Introduction}{

Quality inspection applications in industry are becoming very important. It is a requirement to move towards a zero-defect manufacturing scenario, with unitary non-destructive inspection and traceability of produced parts.
This is one of the applications where image analysis with deep learning (DL) methods is showing its full potential. DL has proven to greatly improve the results of solutions obtained using traditional vision techniques, regarding precision, robustness, and flexibility. These improvements allow models to be adapted  
to incorporate new features of interest, transfer learned models between different domains, and to speed-up the design and development of models for new tasks.

However, the quality inspection environment in the industry has peculiarities that must be taken into account when applying DL-based solutions. Thus, it is not easy to generate large enough data sets with representative images of the different characteristics of interest. Manual labeling of each of the examples must be carried out, which is usually an arduous task that takes large amounts of time and resources. Furthermore, the scarcity of available examples, and the fact that the images of industrial products manufactured successively are very similar to each other, can make the DL algorithm prone to overfitting.
Finally, in new applications or industrial processes, there are no defective data samples from the beginning, so it would be necessary to wait a long time to be able to have DL models capable of identifying the faults that may appear.

Thus, this work presents a methodology to deal with these peculiarities. This methodology should work as a guide towards robust classification and segmentation models, giving rise to fault detection models that are able to detect anomalous feature patterns from the start-up of a new line. The methodology will make use of an anomaly detection model which allows anomalous patterns to be detected in the produced parts, and in addition, the detected anomalies will serve as automatic annotations making the labeling of the images much faster.
The work will be tailored to the solar cell manufacturing industry; however, it could be extrapolated to different domains.

In the last decade, about 2.6 trillion dollars have been invested in renewable energies, half of it in solar energy, with the objective of developing efficient alternatives to traditional energy sources such as oil or gas \citep{frankfurt2019report}. The development of the technology has reduced solar electricity generation cost per kilowatt-hour by 81\%. This cost reduction has turned solar energy into an attractive source of energy for electricity production, increasing the installation of Photovoltaic (PV) cells by 36.8\% between 2010 and 2018 \citep{frankfurt2019report}. This investment trend is expected to continue in the coming years \citep{IEA2019}.

During the assembly of the panels, different events, such as excessive mechanical stress on the panel or a soldering failure, can lead to defects that can harm the long-term energy generation capacity of the module. A defect that covers  8\% of the total cell area may not have a significant impact on the performance if the cell is isolated. However, the same area can have a significant impact when cells are connected and soldered to each other in cell arrays \citep{Koentges2011}, which is the most common layout. The defective area may spread with time, breaking the cell and considerably reducing the energy production capacity of the module. 
As cell production increases, quality inspection becomes critical to avoid defective cells being assembled into the final panel, and thus, to ensure high efficiency and reliable performance of the produced panels.

Nowadays, different imaging techniques are used during PV module inspection to obtain images where defects appear highlighted, for example, Electroluminescence  (EL) \citep{Bartler2018, Chen2020a}, Photoluminescence (PL) \citep{Demant2016, Nos2016}, or Thermography \citep{Pierdicca2018, vanvek2016automation}.
During the assembly stage, EL is one of the predominant techniques. In EL, the cells emit light under electrical current by the phenomenon of Electroluminescence. This light is then captured in high-resolution images where defective areas, with less current flow, appear darker than the remaining parts of the cell \citep{Fuyuki2009}. The most common defects that may arise are cracks, breaks, and finger interruptions \citep{Tang2020}. Figure \ref{fig:types_of_defects} shows the appearance of these kinds of defects in EL images. This technique requires a high degree of control over environmental conditions since the images have to be taken in total darkness. This requirement makes the application of EL unfeasible for outdoor panel inspection, but suitable for inspection in the manufacturing phase with controlled environmental conditions. EL provides high-resolution images where defects are highlighted.

\begin{figure*}[!ht]
    \includegraphics[width=\textwidth]{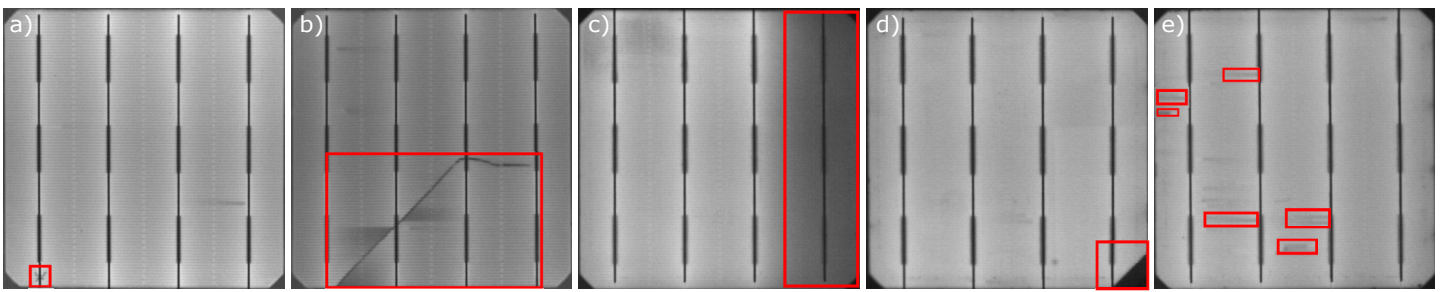}
    \caption{Different type of defects highlighted in Electroluminescence images. The defects are: \textbf{a)} microcrack, \textbf{b)} crack, \textbf{c)} bad soldering, \textbf{d)} break, and \textbf{e)} finger interruptions.}
    \label{fig:types_of_defects}
\end{figure*}

Despite these enhanced images, the defect detection process has to be done by checking each of the cells individually. This process is currently done to a great extent by human operators, who are prone to error, as it is hard for humans to meet industrial production cycle-times. For example, a panel composed of 60 PV modules must be examined in under 30 seconds, which means half a second for each module. Also, human subjectivity is inevitable when deciding if a cell is defective or not, affecting the quality inspection effectiveness. In recent years, several proposals have been made towards the automation of quality inspection. By automating the inspection, all the cells can be checked faster and always using the same objective criteria, overcoming the previous limitations. 

The proposed approaches for automatic PV module inspection can be grouped into three categories according to the required level of human intervention: 1) traditional image processing based approaches, where the procedures used to highlight and binarize defective areas in the images must be manually defined, 2) shallow learning approaches, where machine learning techniques are used for defect identification based on meaningful features that must be obtained through manual feature engineering, and 3) deep learning techniques, where the features are automatically obtained from the data. Note that higher levels of human intervention in the image processing algorithm implementation implies larger development times in order to adapt it to new requirements.

The remainder of the paper is organized as follows: 
Section \ref{sec:related_works} presents some background and related works in the field of photovoltaic cell inspection. 
In Section \ref{sec:methodology}, the unsupervised and supervised training are explained. Section \ref{sec:evaluation} details the dataset, metrics, and hardware and software specifications used in the experiments. Section \ref{sec:experiments}
describes the performed experiments and their results. Finally, Section \ref{sec:conclusions} provides some conclusions about the work.
}

\section{Related works}{
\label{sec:related_works}
In this section, some of the proposed approaches for the automatic detection of defects in images of PV modules are going to be summarized.

The traditional image processing methods are mainly based on manual feature engineering. In this process, the discriminating characteristics of the defects are used to process and binarize the images to highlight the defects. For example, using anisotropic diffusion filters \citep{ko2010anisotropic, Anwar2014} or modified steerable filters \citep{Chen2019, Chen2018e}, the background in the modules is smoothed such that only defects remain. Or inversely, applying anisotropic diffusion \citep{Tsai2010} or filters in the frequency domain \citep{Tsai2012} to remove the defects in the cells, so then the difference between the filtered and the original image is used to highlight the defects.

In other works, the manual feature extraction is combined with shallow learning methods: In \citep{tsai2013} and \citep{Zhang2013}, they extract Independent Component Analysis basis (ICA) from defect-free solar cells samples to construct a demixing matrix. At the inspection stage, the images are reconstructed using the learned basis images and the reconstruction error is used for detecting the presence of defects. In \cite{Rodriguez2021}, 20 different LoG-Gabor Filters are used to extract 81 features for each pixel in the images. Then, Principal Component Analysis (PCA) is used to refine these features, and finally, a Random Forest model classifies each pixel as non-defective or defective. In \citep{Tsai2015}, they extract characteristics of local grains patterns and clusterize them using Fuzzy C-means. At testing, the distance of the grains from the samples to the clusters is used to decide if the grain is defective or not. Similarly, in \citep{Su2019a}, they use a modified Center-Symmetric Local Binary Patterns (CS-LBP) feature descriptor to extract features from the defective areas in the cells, which are then used to train the K-means algorithm. The cluster centroids from training samples are employed to generate global feature vectors to train a classification algorithm, such as a Support Vector Machine (SVM).

Overall, both traditional image processing methods and traditional methods in combination with shallow machine learning techniques can achieve high defect detection rates. However, manual feature engineering is usually time-consuming and requires high domain knowledge. In addition, inspection systems based on these approaches are commonly very case-specific solutions that lack adaptability. A change in the data can mean a substantial change in the inspection system, which would require additional time-consuming manual feature engineering labor to adapt it.

In more recent works, DL methods have been widely applied in the solar cell inspection field. These methods can directly extract meaningful features from the raw data without any feature engineering work, thus making these methods more flexible to changes. The following works  \citep{Bartler2018, Akram2019, Dunderdale2019, Deitsch2018a, AKRAM2020175, Chen2020} are some examples of how Convolutional Neural Networks (CNN) have been employed for classifying solar cells as defective or defect-free during quality inspection. In addition to classification, in some cases, the location of the defects in the cells is also provided. 
For example, in our previous works, we used the sliding window approach with a CNN designed for classification to process cell images by patches and accumulate the results in a heatmap-like image, highlighting areas with a high probability of being defective \citep{balzategui2019}. Or we explicitly train a Fully Convolutional Network to perform pixel-wise classification \citep{balzategui2020defect}. Additionally, other researchers have also proposed other types of defect location, using bounding boxes \citep{Liu2019a, Zhang2020}, or by visualizing the activation maps from the last network layer \citep{Deitsch2018a, mayr2019}.

Nevertheless, to obtain high detection rates, the networks are trained using supervised learning, which requires a considerable amount of annotated defective data. The quality of the results (i.e., detection rate) in supervised learning is directly proportional to the amount of employed annotated data. However, this represents a challenge in many industrial applications as sufficient defective samples may be difficult to obtain in an industrial setting.
Thus, the creation of accurate inspection models may be difficult, as a new manufacturing line will need time to generate a representative dataset with enough examples. There may also be certain very rare defect types that might be difficult to gather for the dataset.

To tackle the problem of insufficient defective data, several researchers have proposed different solutions. One of the approaches is transfer learning \citep{AKRAM2020175, demirci2019defective, Qian2020a}, where the neural network is initialized using weights from a previously trained network. Then, the model is refined using a few case specific samples. Transfer learning is limited by the similarity between the source and target domains. Currently, the available pre-trained weights have been mainly trained on natural images rather than on industrial datasets, which can limit their use in industrial cases.

Another approach consists in generating synthetic data to compensate imbalanced datasets employing variants of the Generative Adversarial Network (GAN) \citep{Goodfellow2014}. These architectures have shown remarkable capabilities in learning latent representations of real data to generate realistic synthetic samples. In this way, synthetic defective samples are generated and employed along with real samples to train a conventional CNN. This approach alleviates the risk of overfitting and improves the generalization capability of the network. This strategy has been successfully employed to generate realistic human faces \cite{choi2018stargan}, synthetic machinery faulty signals \cite{ZHANG2020107377}, and also defective solar cell samples \citep{Tang2020, Luo2019}. Nonetheless, both Transfer Learning and GANs still require defective data.

In other domains, researchers have used an anomaly detection approach to avoid the need for defective data. The objective of this approach is to train a network to learn the probabilistic distribution of normal data. Then, the learned features can be used to discriminate samples that will be far from what is considered normal, and thus, detect defective samples. In anomaly detection, just defect-free samples are used during training and there is no need for annotations. These features make anomaly detection an interesting approach for industrial applications. Anomaly detection has been applied in different industrial cases, e.g., \citep{Haselmann2018, Staar2019} and also in the medical domain, e.g., \citep{Schlegl2019, Chen2018a}, where it is also difficult to obtain anomalous data for training. However, these approaches usually result in less accurate models than those obtained with supervised training.

In the case of solar cell inspection, anomaly detection approaches have been proposed in \citep{Qian2020} and \citep{Qian2020a}, where they train a Stacked Denoising AutoEncoder (SDAE) to extract features from defect-free samples using the sliding window method. In \citep{Qian2020a}, they extend the network architecture with a pre-trained VGG16 network that works as an additional feature extractor. This extra branch extracts additional features which are  fused with the already extracted feature maps enhancing the obtained information.
At testing time, the same procedure is applied and the extracted features are processed using matrix decomposition to localize the defects in the cells. After that, some morphological processing is applied to improve the results. However, in these works, only the detection of cracks is targeted. Furthermore, the images are processed using the sliding window method. This procedure slows down the inspection process limiting its deployment into a real production environment.

An inspection system should be able to detect the maximum number of defects, be fast to meet the established inspection time, and require the minimum human intervention in order to save resources and time. The main contribution of this work is a methodology that tries to meet these requirements by combining the accuracy of supervised models with the benefits of an anomaly detection approach, i.e., that it only requires defect-free samples for training and avoids the need for data labeling. The approach has been tailored for the detection and segmentation of different types of defects such as cracks, microcracks, or finger interruptions in EL images of solar cells; however, it should also be applicable in other industrial inspection tasks. The methodology is illustrated in Figure \ref{fig:whole_training} and consists of two stages:

\begin{figure*}[!ht]
    \centering
    \includegraphics[width=\textwidth]{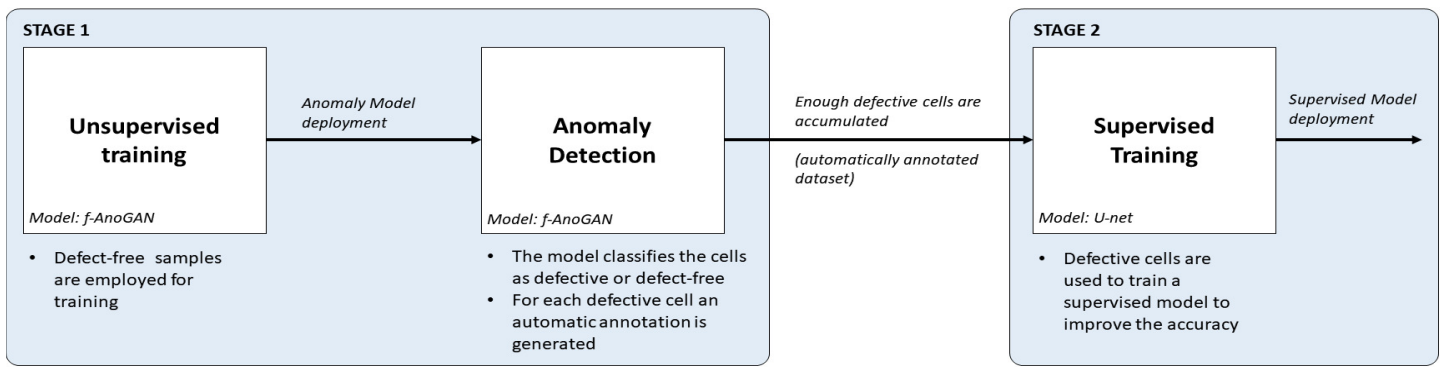}
    \caption{General schema of the proposed methodology. 1) In the unsupervised training, the network for anomaly detection is trained. 2) In anomaly detection, the network from the previous step is applied to detect and locate defects, and thus, generate an automatically labeled dataset. 3) Finally, the generated dataset is employed to train a supervised model that improves the detection rates.}
    \label{fig:whole_training}
\end{figure*}

\begin{enumerate}
    \item First, using an anomaly detection approach, defect-free samples can be employed to obtain an initial inspection model that from the very beginning of a new production line can detect and segment anomalies in EL images of cells. For this purpose, f-AnoGAN \citep{Schlegl2019}, a GAN based anomaly detection network that has been shown to work well with medical images, is adapted for inspection. The original architecture has been modified such that instead of using a sliding window method, the images can be processed as a whole, reducing the processing time drastically. Also, a modified training scheme is proposed which improves the defect detection rates with respect to the results with the original training scheme.
    
    \item Then, as defective cells arise, the anomaly detection model will separate them from the defect-free ones and it will generate pixel-level annotations without any human intervention. The experiments have shown that these segmentation results can be used as pixel-wise labels for the supervised training of a U-Net \cite{DBLP:journals/corr/RonnebergerFB15} based model that improves the defect detection rates of the anomaly detection model.
\end{enumerate}

}

\section{Methodology}{
\label{sec:methodology}
This section details how the different networks used in the methodology are trained.

\subsection{Unsupervised Model for Anomaly Detection}{

In this stage, the objective is to train an anomaly detection model that can detect and locate anomalous patterns within solar cell images. This is achieved by training f-AnoGAN network to encode and reconstruct only defect-free samples, so then, when processing defective samples, it will output a defect-free version of them. The differences between the original and the reconstructed defect-free version will highlight anomalies in the cells.

f-AnoGAN is composed of three different sub-networks (a generator $G$, a discriminator $D$, and an encoder $E$) that are trained in two phases.

In the first training phase, the generator and discriminator are trained in an adversarial manner to learn a latent space of normal data variability using just normal data. In this work, defect-free samples are considered as normal data and defective samples as anomalous data. 

In the second phase, the encoder is trained to map normal data from the image space to the learned latent space while the Generator and Discriminator are kept unaltered. Once these two phases have finished, the encoder can map test images from the image space to the latent space, and the generator can reconstruct the encoded version of the images from the latent space back to the image space. As the network is trained on normal data, it only learns to encode and reconstruct correctly normal features, thus when processing anomalous samples, deviations from the reconstructed images can be used for anomaly detection and location. 

\subsubsection{Phase 1 - WGAN Training}{
The objective of the first training phase consists in learning the variability of normal data. For this purpose, a Wasserstein GAN (WGAN), composed of a generator and a discriminator, is optimized to learn the normal data probability distribution. The optimization is achieved using the gradient penalty based loss shown in Equation \ref{eq:w_loss} proposed by \citep{Gulrajani2017}, where the Wasserstein distance between the real normal data probability distribution $P_r$, and generator synthesized data probability distribution $P_g$ is minimized.

\begin{equation}
\label{eq:w_loss}
    L_{WGAN} = \underset{\Tilde{\mathbf{x}}\sim\mathbb{P}_g}{\mathrm{\mathbb{E}}} [D (\tilde{\mathbf{x}})] - \underset{\mathbf{x}\sim\mathbb{P}_r}{\mathrm{\mathbb{E}}} [D (\mathbf{x})] + \lambda \underset{\hat{\mathbf{x}}\sim\mathbb{P}_{\hat{\mathbf{x}}}}{\mathrm{\mathbb{E}}} [{(\lVert{\nabla_{\hat{\mathbf{x}}}D \hat{\mathbf{x}}) \rVert}_{2}-1)}^2],    
\end{equation}
where $\tilde{\mathbf{x}} = G(\mathbf{z})$, $\hat{\mathbf{x}} = \alpha \mathbf{x}+(1-\alpha)\tilde{\mathbf{x}}$ with $\alpha \sim U(0, 1)$ and $\lambda$ is the penalty coefficient.
During training, the generator is fed with a noise input vector $\mathbf{z}$, sampled from a latent space $\mathcal{Z}$, and tries to learn the mapping from that latent space to the image space $\mathcal{X}$. The synthesized data $G(\mathbf{z})$ should follow as closely as possible the real data distribution $P_r$. Simultaneously, the discriminator is given the generated sample $\tilde{\mathbf{x}}$ and the real sample $\mathbf{x}$ so it outputs a scalar measure of how close both distributions are. The training and the components in this phase are illustrated in Figure \mbox{\ref{fig:gan_train_1}}.

\begin{figure}[!ht]
    \centering
    \includegraphics[width=0.5\textwidth]{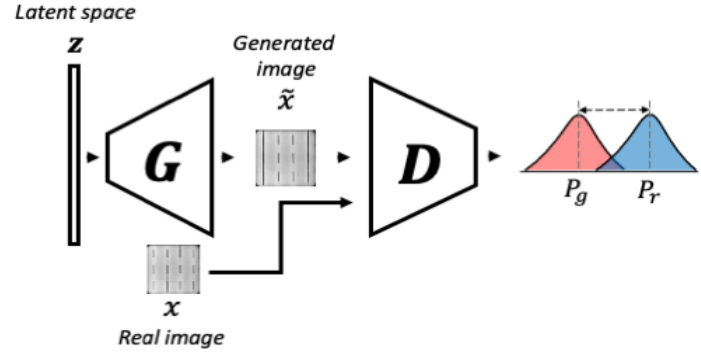}
    \caption{The schema of phase 1 of f-AnoGAN training. The Generator takes a vector $\mathbf{z}$ and tries to generate an image that follows the same distribution of the real data. Then, the Discriminator measures the difference between the generated data distribution and the real data distribution. This schema was inspired by the images from \cite{Schlegl2019}.}
    \label{fig:gan_train_1}
\end{figure}

After the first phase of training, 1) a latent space that represents the variability of the normal data, 2) a generator that can map samples from this latent space to image space, and 3) a discriminator that can detect samples that do not follow the normal data distribution are obtained.

However, at this phase, there is no network component that can perform the inverse mapping, i.e., from image space to latent space. The next phase will focus on learning this mapping.

}

\subsubsection{Phase 2 - Encoder Training}{
\label{subsec:encoder}
In the second training phase, illustrated in Figure \ref{fig:gan_train_2}, the objective is to make the encoder learn to map a real image to the latent space such that the generator can map it back to the image space. During this phase, both the generator's and the discriminator's weights remain unaltered. This network configuration is denoted as $izi$ in \citep{Schlegl2019}. In this case, the encoder is optimized by minimizing the Mean Square Error (MSE) with respect to the difference between the original image $\mathbf{x}$ and the reconstructed one $G(E(\mathbf{x}))$. Additionally, the reconstruction error from the $izi$ architecture loss is extended by including feature residuals from an intermediate layer in the discriminator, yielding the $izi_f$ architecture. 
By taking into account these residuals in the feature space, the reconstruction is improved \citep{Schlegl2019}.
The loss function of $izi_f$ is defined by Equation \ref{eq:enco_loss}:
\begin{equation}
    \label{eq:enco_loss}
    L_{izi_{f}} = \frac{1}{n}\lVert \mathbf{x} - G (E(\mathbf{x}))\rVert_{2} + \frac{k}{n_{d}} \lVert f(\mathbf{x}) - f (G(E(\mathbf{x})))\rVert_{2},
\end{equation}
where $f(\cdot)$ corresponds to the discriminator's intermediate layer features, $n_{d}$ is the dimensionality of the intermediate feature representation, and $k$ is a weighting factor.

\begin{figure}[!ht]
    \centering
    \includegraphics[width=0.5\textwidth]{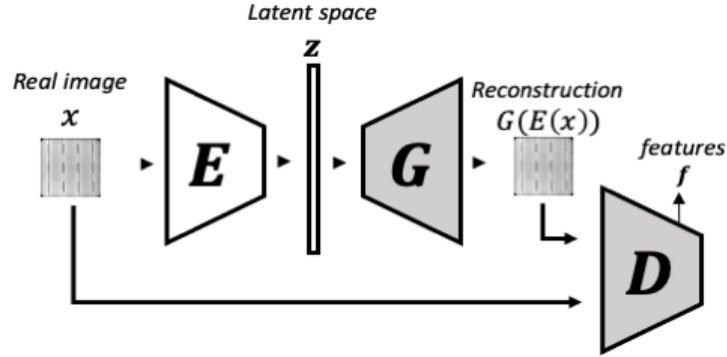}
    \caption{The schema of phase 2 of f-AnoGAN training. In the second phase, the Generator and Discriminator are kept unaltered while an Encoder is added and trained to learn to encode the images to the latent space so the Generator can reconstructed them back. This schema was inspired by the images from \cite{Schlegl2019}.}
    \label{fig:gan_train_2}
\end{figure}

}

\subsubsection{Anomaly Detection}{
Once the training has finished, all the components are fixed and ready to be used for anomaly detection. At this point, the images are processed as in the encoder training. First, the encoder maps the images to the latent space, and then, the generator maps them back to the image space. Finally, the difference between the reconstructed and the original image defined in Equation \ref{eq:anomaly_score} is used for anomaly detection.

\begin{equation}
    \label{eq:anomaly_score}
    A(\mathbf{x}) = A_R(\mathbf{x}) + k \cdot A_D(\mathbf{x}),
\end{equation}
where $A_R(\mathbf{x}) = \frac{1}{n}\lVert \mathbf{x} - G (E(\mathbf{x}))\rVert_{2}$,  $A_D(\mathbf{x}) = \frac{1}{n_{d}} \lVert f(\mathbf{x}) - f (G(E(\mathbf{x})))\rVert_{2}$ and $k$ is a weighting factor from Equation \ref{eq:enco_loss}.

Only defect-free cell samples have been used for training, therefore the network will have only learned to reconstruct normal samples. In the case of defect-free samples, the network outputs an image similar to the input image, thus there is not much deviation when subtracting one image from the other. Instead, when processing a defective cell, the output is a defect-free version of the input sample. As a consequence, the deviation between the original and reconstructed images can be used to detect anomalous parts. This behavior is shown in Figure \ref{fig:fanogan_application}.

\begin{figure}[!ht]
    \centering
    \includegraphics[width=0.65\columnwidth]{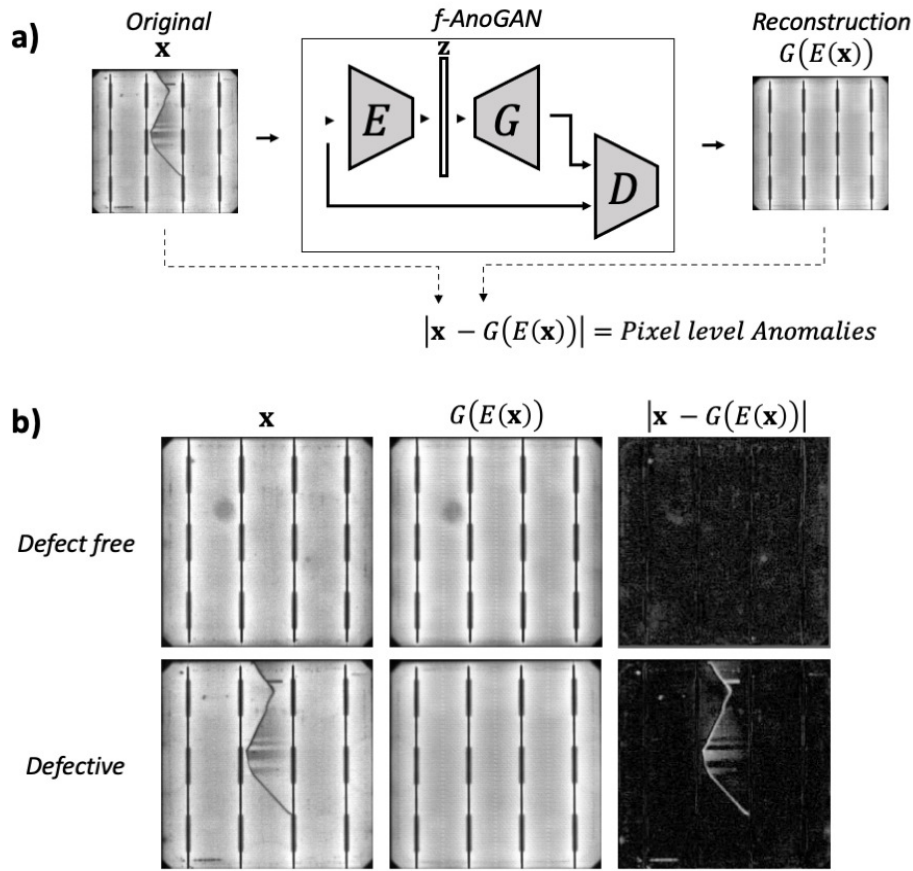}
    \caption{Example anomaly detection with f-AnoGAN. In \textbf{a)} the final structure of the network
used for anomaly detection, and in \textbf{b)} some example results obtained when the network process a
defect-free cell and a defective cell. This schema was inspired by the images from \cite{Schlegl2019}.}
    \label{fig:fanogan_application}
\end{figure}

The absolute value of the pixel-wise difference between the original and the reconstructed image, $\lvert \mathbf{x} - G(E(\mathbf{x})) \rvert$, is used for pixel-wise anomaly detection. By applying a threshold $c$, defined in Equation \ref{eq:thresh_loss}, to the residuals image obtained from $\lvert \mathbf{x} - G(E(\mathbf{x})) \rvert$, the binary image $\mathbf{y} \in \{0,1\}$ is obtained. 

\begin{equation}
    \label{eq:thresh_loss}
    \mathbf{y} = 
    \begin{cases}
            1, & \lvert \mathbf{x} - G(E(\mathbf{x})) \rvert \geq c \\
            0, &otherwise.
    \end{cases}
\end{equation}

This binary image can be considered as a pixel-wise annotation of defective samples as described in Section \ref{sec:supervised_train}.

In this work, two modifications have been made to the original f-AnoGAN network to adapt it for anomaly detection in photovoltaic cell manufacturing. 

With f-AnoGAN, the images are processed in patches of size 64x64 pixels, which requires multiple executions of the network, increasing the time to process an entire cell. As a consequence, the network does not meet the industrial production cycle time (under half a second per cell). In order to reduce the inspection time, the encoder input and the generator output layers' dimension was increased. Thus, whole cell images will be processed in a single pass, reducing processing time drastically with respect to the original sliding window approach.

In addition, the training scheme was also modified. In f-AnoGAN, the generator is frozen during the second training phase in Section \ref{subsec:encoder}, thus, only encoder weights are modified. This can limit the network capability in terms of reconstructing the input image. 
In order to maintain a stable training without restricting the reconstruction capability, the generator is also trained at a certain number of the encoder training iterations with a lower learning rate, while keeping the discriminator unaltered. By training the generator, the reconstruction of defect-free samples will improve. Therefore, the deviation between the original and the reconstructed images of normal data will be reduced. Consequently, both the anomaly score and the pixel differences will be lower for defect-free samples, but higher for defective ones, thus the model's detection rate will improve.
}

}
\subsection{Supervised Model for Defect Segmentation}{
\label{sec:supervised_train}

In anomaly detection, the model is taught to find everything that is not considered normal. 
In supervised training, the model is instead trained with labels to search for specific defective patterns in the data, which usually yields more precise models for defect detection. Using the anomaly detection approach as an automatic labeling method, one may benefit from the precision of supervised learning models avoiding the time-consuming, and not always trivial, pixel-level labeling task, thereby considerably reducing the effort dedicated to the setup of a new inspection system. 

This way, in the first stage of the inspection system development, where lots of defect-free cell samples and few defective cell samples are available, an initial inspection model can be obtained using anomaly detection. Then, as defective cells arise, the trained anomaly model will process the samples and output pixel-wise annotations avoiding the time-consuming data annotation task. After some time, when there are enough annotated defective cell samples, a model will be trained in a supervised manner to search for specific features in the images as in our previous works \citep{balzategui2019, balzategui2020defect}, obtaining more accurate models. 

For the supervised training, as in our previous work \citep{balzategui2020defect}, U-net \citep{DBLP:journals/corr/RonnebergerFB15}, an end-to-end trainable Fully Convolutional Neural Network (FCN), was used. This network has been shown to work well on biomedical image segmentation with low amounts of data. The network follows an encoder-decoder shape where, after successive downsampling and then upsampling steps, features from the images are extracted to finally output a segmentation map of the same size of the input. Additionally, skip connections connect blocks in the encoder and decoder parts helping to recover fine-grained details lost during the downsampling and improving the final results.
}

}

\section{Experimental setup}{
\label{sec:evaluation}

To validate the proposed methodology, several experiments were carried out regarding unsupervised training and supervised training. In this section, the characteristics of the employed industrial dataset, the metrics used to evaluate the performance of the models, and the hardware and software specifications will be described

\subsection{Dataset}{
\label{subsec:dataset}

The employed dataset is composed of Electroluminescence images of 1873 monocrystalline solar cells extracted from 31 panels. The images were provided by Mondragon Assembly S. Coop. and were taken at the quality inspection stage during the assembly of the solar panels. The cells have a size of 15x15 cm and the images have an average resolution of 840x840 pixels.

The dataset is composed of 1498 images of cells considered defect-free by the company and 375 defective cells containing cracks, microcracks, and finger interruptions. The distribution of the dataset is shown in Table \ref{tab:dataset}. Each defective sample has its manually labeled pixel-wise binary annotation $\{0,1\}$ as shown in Figure \ref{fig:dataset_sample}.

\begin{specialtable}[H]
\caption{Dataset sample distribution.}

\centering
\begin{tabular}{lcc}
\toprule

\multicolumn{1}{c}{\textbf{}} &  & \textbf{Total} \\ \midrule
\multicolumn{2}{l}{\textbf{Defect-free}} & \textbf{1498} \\
\multicolumn{2}{l}{\textbf{Defective}} & \textbf{375} \\
\multicolumn{2}{r}{Crack} & 18 \\
\multicolumn{2}{r}{Microcrack} & 240 \\
\multicolumn{2}{r}{Finger interruptions} & 117 \\ \bottomrule
\end{tabular}

\label{tab:dataset}
\end{specialtable}

\begin{figure}[h]
    \centering
    \includegraphics[scale=0.7]{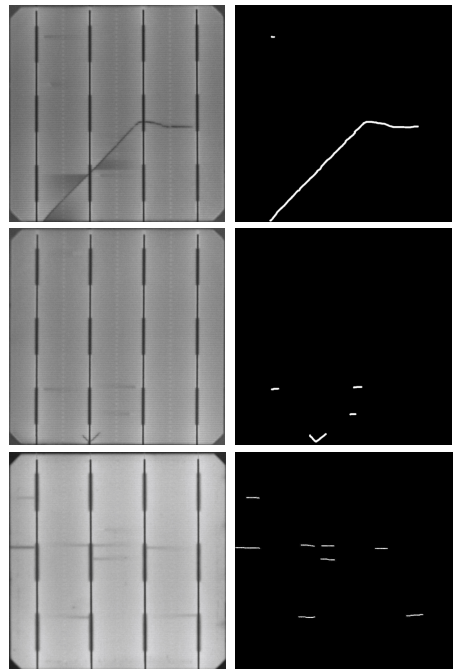}
    \caption{Defective monocrystalline samples with their pixel-level annotations.}
    \label{fig:dataset_sample}
\end{figure}

}

\subsection{Metrics}{

The results from the experiments were quantitatively and qualitatively measured. For the defect detection performance assessment in the unsupervised part experiments, the anomaly score from Equation \ref{eq:anomaly_score} was used to construct the Receiver Operating Characteristic (ROC) curves and to calculate the Area Under Curve (AUC). 
The results were analyzed taking into account all the defects as if they were from the same class, and also considering each defect type separately.

Note that there is a great unbalance between samples from each defective class and defect-free samples as shown in Table \ref{tab:dataset}. So, to avoid misleading conclusions when interpreting the ROC curves, the data was balanced by first taking all defective samples of the class that was being analyzed, and then, randomly selecting the same amount of defect-free samples.

Then, the results were binarized to calculate Precision, Recall, Specificity, and F1-score metrics, defined in Equations \ref{eq:precision}, \ref{eq:recall}, \ref{eq:specificity}, and \ref{eq:f1} respectively.

\begin{equation}
    Precision = \frac{TP}{TP + FP}
    \label{eq:precision}
\end{equation}
\begin{equation}
    Recall = \frac{TP}{TP + FN}
    \label{eq:recall}
\end{equation}
\begin{equation}
    Specificity = \frac{TN}{TN + FP}
    \label{eq:specificity}
\end{equation}
\begin{equation}
    F1\_score = 2\cdot\frac{Precision \cdot Recall}{Precision + Recall}
    \label{eq:f1}
\end{equation}
where TP stands for True Positive, TN for True Negative, FN for False Negative, and FP for False Positive. In this work, defective samples belong to the Positive class and defect-free samples to the Negative class.
}

As there were few defective samples left for testing the supervised part, the performance of the models was mainly evaluated at a qualitative level, although quantitative analysis has also been reported at image-level. The Recall, Precision, and Specificity metrics have been applied considering the cells as defective if they contained more than 20 defective pixels.

\subsection{Hardware and software}{
Two Nvidia GeForce RTX 2080 GPUs were used in the experiments. The models' training in the unsupervised part required both GPUs, while the anomaly detection and supervised training only required one GPU. For all f-AnoGAN based models, the publicly available code \footnote{\url{https://github.com/tSchlegl/f-AnoGAN}} was used, which is written in Python2.7 and employs Tensorflow 1.2 and CUDA 8. Instead, in the Convolutional Deep Autoencoder and supervised U-net training, Python 3.6, Tensorflow 1.14, and CUDA 10 were used.}
}
\section{Experiments}{
\label{sec:experiments}
In this section, the experiments and the results from each part (unsupervised and supervised) are going to be described.

\subsection{Unsupervised Model for Anomaly Detection}{

In this part, the impact of the proposed network modifications with regards to the processing time and defect detection rates is evaluated. First, the technical details about the experiments are going to be explained, and then, the results obtained from the experiments will be analyzed and compared.

\subsubsection{Experimental Design}{

First, the network in \citep{Schlegl2019} was applied in our dataset to ensure its applicability in this specific industrial context. Then, the modifications regarding the input size and training scheme were incorporated. These two models are referred to as f-AnoGAN-64 for the original network configuration, and f-AnoGAN-256 for the model with the modifications.

The hyperparameters in both models were all kept the same as in \citep{Schlegl2019}: The $\mathbf{z}$ vector was sampled from a Normal distribution and had a size of 128, the value of $\lambda$ parameter for the gradient penalty was 10, and the value for the weighting factor $k$ in Equation \ref{eq:enco_loss} was set to 1. The optimization algorithm for the first training phase was Adam \citep{kingma2014adam} and for the second RMSprop \citep{hinton2012neural}. For both models, all the images were rescaled to a range [-1,1] as in the original work \mbox{\citep{Schlegl2019}}. In this way, the pixels in the images will match the range of the Generator output layer activation function (i.e., tanh), and it will help the network have a stable training \mbox{\citep{chintala2016train}}. The only hyperparameter that was modified was the batch size for f-AnoGAN256 training in order to fit the model in memory. This was due to the increase in trainable parameters resulting from the modification of the architecture. The training took a different number of iterations depending on the phase. Both models required 40k iterations in the first phase and 70k iterations in the second phase to converge.

In addition to the mentioned models, two Convolutional Deep Autoencoders were also trained. These models were used to establish a base with which the results from the previous two models could be compared. In addition, the results from these two base models served to check if a more simple network architecture could be enough to obtain high defect detection rates in this specific context. Following the two approaches from previous models, one Autoencoder was trained to process the images in patches, and the other Autoencoder was trained to process the images in an image-wise setup. These models will be referred to as AE-64 and AE-256.  

Regarding the architectures, both networks are composed of an encoder and a decoder with several convolutional layers. In the case of AE-64, the encoder has two convolutional layers with 64-32 filter distribution, followed by 4 Fully Connected layers of 128 units each, and finally a decoder with the inverted shape of the encoder part. In the case of AE-256, the architecture is two convolutional layers deeper than the AE-64 such that the output dimension before the Fully Connected layers is the same. The filter distribution is 8-16-32-64. After each Fully Connected layer, a dropout layer with a drop rate of 0.25 was set. Both networks were optimized with the MSE loss function and Adam as the optimization algorithm. The AE-64 model training took about 30k iterations with a batch size of 32, and the AE-256 model training took about 6k iterations with a batch size of 8. 

In the experiments in this part, only defect-free cell images were required for training. The defect-free samples in Table \ref{tab:dataset} were separated into the train, validation, and test sets. In addition, the test set also contained 375 defective samples. The dataset distribution used in this part is illustrated in Table \ref{tab:dataset_unsupervised}.

\begin{specialtable}[H]
\caption{Dataset sample distribution for unsupervised part experiments.}
\centering
\begin{tabular}{lccccc}
\toprule
\multicolumn{1}{c}{\textbf{}} &  & \textbf{Train} & \textbf{Val} & \textbf{Test} & \textbf{Total} \\ \hline
\multicolumn{2}{l}{\textbf{Defect-free}} & 750 & 373 & 375 & 1498 \\ 
\multicolumn{2}{l}{\textbf{Defective}} & - & - & 375 & 375 \\
\multicolumn{2}{r}{Crack} & - & - & 18 & - \\ 
\multicolumn{2}{r}{Microcrack} & - & - & 240 & - \\ 
\multicolumn{2}{r}{Finger inter.} & - & - & 117 & - \\ \bottomrule
\end{tabular}

\label{tab:dataset_unsupervised}
\end{specialtable}

f-AnoGAN-64 and AE-64 were designed to process the images patches-wise. For these cases, each image was split into 256 patches using a sliding window. The final train, validation, and test sets were composed of 192000, 95488, and 192000 images, respectively. For the other networks, the images were resized to the network input size (i.e., 256x256 pixel resolution).

To compare the performance between the models, the results from the patch-based models were post-processed. While in the image-wise models it was enough to apply a single threshold so as to classify a sample as defective or non-defective, in patch-based models the errors of all patches belonging to the cell must be taken into account. So, in the latter, the same threshold was applied to every patch and if a single patch was evaluated as defective, the entire cell was also evaluated as defective.

}
\subsubsection{Results}{

Regarding the image-level results, the network modifications had a positive impact on the results. f-AnoGAN-256 was able to detect more defects (higher Recall values) than the original f-AnoGAN-64 without incurring in more False Positive cases (higher Precision and Specificity values). This is particularly visible in Table \ref{tab:results_img} for the case of the finger interruption and microcrack defect classes where all the metrics improved over 10 points. This improvement can also be appreciated in the ROC curves and the AUC values in Figure \ref{fig:Results_ROC} where the curve reflecting the performance of f-AnoGAN-256 appears closer to the top left corner that represents the perfect classifier, and the AUC value that changed accordingly.

\begin{figure*}[!h]
    \centering
    \includegraphics[width=0.9\textwidth]{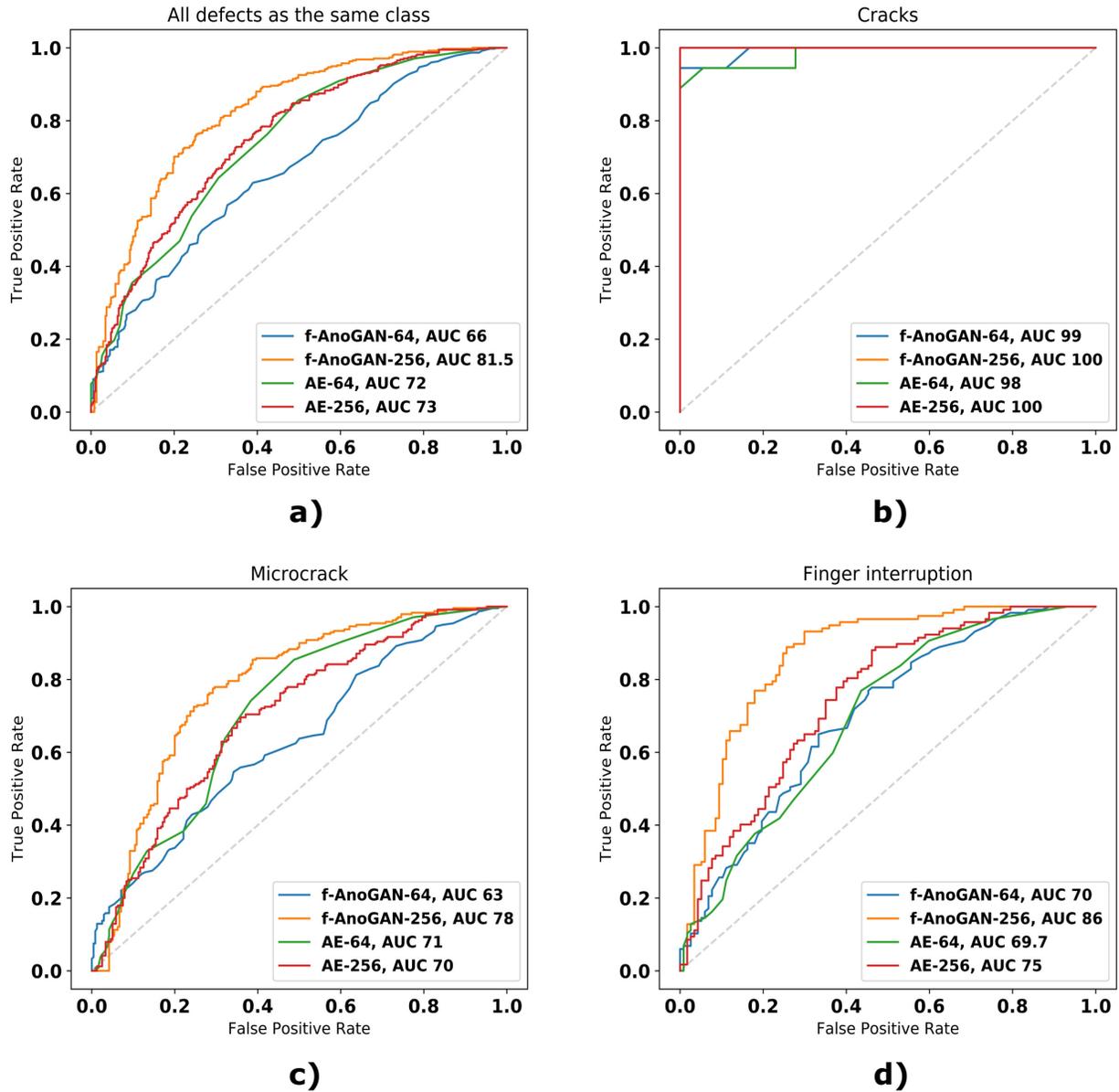}
    \caption{ROC curves from the different models in the unsupervised part experiments. \textbf{a)} Results
considering all defects as they belong to the same class, \textbf{b)} results with defective samples with cracks,
\textbf{c)} results with defective samples with microcracks, \textbf{d)} results with defective samples with finger
interruptions.}
    \label{fig:Results_ROC}
\end{figure*}

\begin{specialtable}[!ht]
\centering
\caption{The results of anomaly detection at the image-level. Precision tells how accurate the classifier is when classifying a sample as defective. Recall tells how many samples have been correctly classified as defective from all defective samples. Specificity describes how many defect-free samples have been correctly classified as defect-free samples. The F1-score is the harmonic mean of the Precision and Recall. In all metrics, the higher the value, the better the classifier is.}

\begin{tabular}{llcccccc}
\toprule
    &    \textbf{Model}&    \textbf{AUC}&    \textbf{Precision}&     \textbf{Recall}&      \textbf{Specificity}&    \textbf{f1-score}\\ \midrule
    All test samples &     &      &      &    &            &      \\
    &    f-AnoGAN-64   &     66   &    61.3 &    62.8&         61 &     62  \\
     &    f-AnoGAN-256&     \textbf{81.5} &    \textbf{75} &     \textbf{78}&    \textbf{75}&     \textbf{77}  \\
    &    AE-64         &     72 &    65.6 &     64&        68 &     65  \\ 
    &    AE-256&    73&    68.4&    58&       72&    63\\\midrule
    
    Cracks&     &      &      &            &      \\
    &    f-AnoGAN-64&     99 &     66.7 &     100&         50 &     80 \\
    &    f-AnoGAN-256&     \textbf{100} &    \textbf{95}  &     100 &        \textbf{94} &     \textbf{97} \\
    &    AE-64&     98 &     78   &     100 &    100&        87.7  \\
    &    AE-256&    100&    95&    100&       94&    97\\\midrule
    
    Micro&     &      &      &      &         &      \\
    &    f-AnoGAN-64 &     63 &     58.7  &     59&         59 &     58.9 \\
    &    f-AnoGAN-256&     \textbf{78} &     \textbf{73}&     \textbf{73} &        \textbf{74 } &     \textbf{73} \\
    &    AE-64&     71 &      66.5  &     63.7 &       67.9 &     65  \\
    &    AE-256&    70&    66&    53&       72&    59\\\midrule
    
    Finger int.&     &      &      &           &     &      \\
    &    f-AnoGAN-64 &     70 &     66 &     64.9 &        66.7&     65.5 \\
    &    f-AnoGAN-256&     \textbf{86} &    \textbf{ 78}  &    \textbf{85} &        \textbf{75} &    \textbf{81} \\
    &    AE-64&     69.7 &     61.9   &     59.8 &       63 &     60.8  \\
    &    AE-256&    75&    69&    63&       71&    66\\
    \bottomrule
\end{tabular}
\label{tab:results_img}
\end{specialtable}

If the results of f-AnoGAN models are compared with the ones from the Autoencoders, it is further underlined that the incorporated of network modifications brought an improvement in defect detection rates. Setting aside the case of finger interruptions, f-AnoGAN-64 obtained worse detection rates than its Autoencoder counterpart (i.e., AE-64) and also AE-256. But, when the proposed changes were incorporated, the obtained results surpassed the ones from the Autoencoders for all the classes and all the metrics, which means higher True Positives cases and lower False Positive cases for all defect classes.

The results in the ROC curves in Figure \ref{fig:Results_ROC} and the metrics in Table \ref{tab:results_img} show that all models could detect all samples with cracks, but they could not detect all samples with microcracks and finger interruptions. This is caused by the fact that cracks are defects that cover a larger area of the cells than finger interruptions or microcracks, and therefore have more defective pixels that result in a higher anomaly score. The same happens in the case of finger interruptions and microcracks. The first appear in groups of three or more, whereas the latter appears isolated. Because of this, the sum of defective pixels in samples with finger interruptions contributes to higher anomaly scores resulting in higher detection rates.

With respect to the defect location results, Figure \ref{fig:Results_pixel_level} shows that all the models were able to properly locate the different defect classes. Nevertheless, the segmentation results were more refined in f-AnoGAN-256 and AE-256 models. Although the patch-based models were able to point out the presence of defective areas, the borders and shape of the predictions were not as accurate as the ones from the image-wise models.

Also, the patch-based models have more False Positive cases. An example of this behavior is the sample from the second row, where the buses in the cell were mistakenly detected as defects. The same happened on the defect-free samples, where patch-based models classified defect-free areas as defective (e.g. samples seven and eight), whereas the image-wise models obtained clean predictions. Although not illustrated in Figure \ref{fig:Results_pixel_level}, this behavior was shared across several other samples in the test set.

\begin{figure*}[!h]
    \centering
    \includegraphics[width=0.75\textwidth]{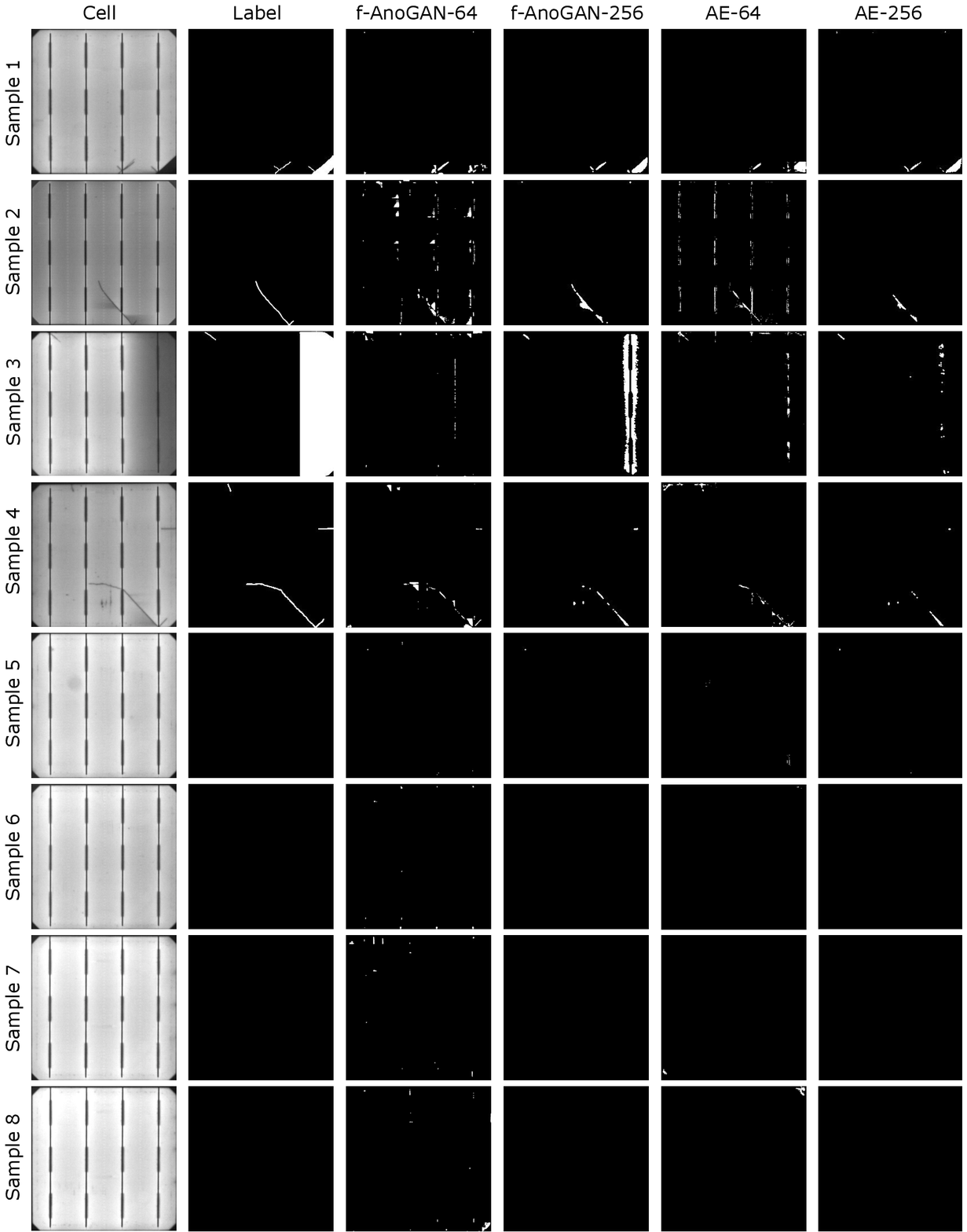}
    \caption{Defect localization results from each model.}
    \label{fig:Results_pixel_level}
\end{figure*}

In addition to the classes in quantitative analysis, the models were also executed on samples that contained two other defects that were put aside as they were few available samples for a proper analysis. These defects classes were breaks and bad soldering present in Figure \ref{fig:Results_pixel_level} in the bottom-right of the first row sample and on the right in the third row sample, respectively. In the case of the break, image-wise models were able to output a relatively precise segmentation. The AE-64 model results indicated the defect location, however, they did not have much precision. Instead, in the case of f-AnoGAN-64, it can be noticed that at the defect location there is a certain anomalous pattern but very vaguely segmented. Regarding the bad soldering, f-AnoGAN-256 was the only model that presented a reasonable segmentation result.

Regarding the processing time, Table \ref{tab:exec_time} shows how the architecture modification made f-AnoGAN able to reduce the time required to process each cell. While patch-based models required more than half a second to process the cells (maximum stipulated time per cell), f-AnoGAN-256 and AE-256 required only 0.05 and 0.02 seconds respectively to process each cell. 

\begin{specialtable}[H]
\caption{Time required to process a cell for each model.}
\centering
\begin{tabular}{lcc}
\toprule
        \textbf{Model}&    \textbf{Time per patch}&    \textbf{Time per image}\\ \midrule
        f-AnoGAN-64  &   0.02s   &    5.12s\\
        f-AnoGAN-256  &   -   &    \textbf{0.05s} \\
        AE-64         &   0.012s  &    3.07s   \\ 
        AE-256&    -&    \textbf{0.02s}  \\
    \bottomrule
\end{tabular}

\label{tab:exec_time}
\end{specialtable}

}

Overall, it can be concluded that the proposed modifications made the network obtain higher defect detection rates and also reduced the processing time meeting the established time for industrial inspection.
}

The results in this part have shown that the anomaly model can yield relatively high detection rates and defect locations. Among the trained models, f-AnoGAN-256 has shown the highest detection rates, short enough processing time for industrial inspection, and precise pixel-level results. Taking into account that the defect location results in Figure \ref{fig:Results_pixel_level} were close enough to what human experts annotated, this model was selected as the automatic annotator model to obtain automatic labels for the next stage in the methodology (i.e., supervised stage).
}

\subsection{Supervised Model for Defect Segmentation}{
The goal of this experiment was to prove the feasibility of using the anomaly detection approach as an automatic labeling method. To this end, the segmentation results from a model trained on automatic labels obtained from the previous stage and a model trained with labels created by experts were compared. Some samples of the automatic labeling used for training are shown side by side with their corresponding manually labeled counterpart in Figure \ref{fig:automatic_label}.

\begin{figure}[!ht]
    \centering
    \includegraphics[width=0.4\columnwidth]{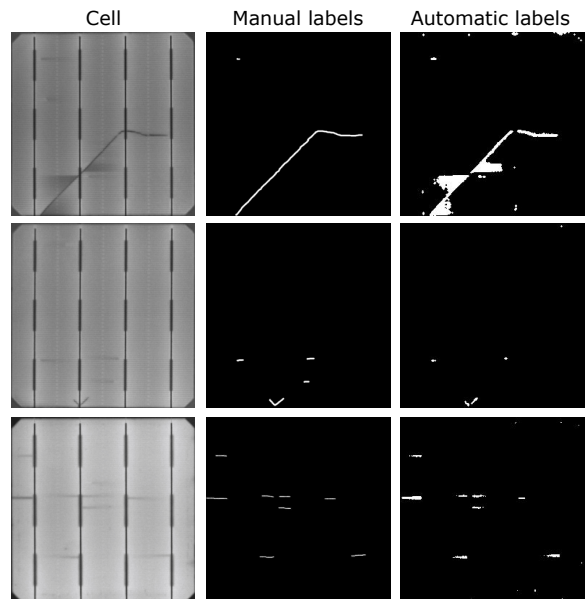}
    \caption{Manual and automatic labeling for different samples. The automatic labeling kept the segmentation of the labeled defects, but at the same time introduced some additional areas. This is especially noticeable in the samples from the first row, where the manual labeling only considered the defect itself, but the automatic labeling also considered the darker areas created by the effect of the defect.}
    \label{fig:automatic_label}
\end{figure}

In the same way as in the previous section, first the technical details of the experiments are going to be described, and then, the results are going to be analyzed.

\subsubsection{Experimental Design}{

For both manual and automatic labeling models, U-net was used. The network configuration was kept as in \citep{DBLP:journals/corr/RonnebergerFB15}, which is composed of 9 convolutional layers blocks: 4 in the encoder, 4 in the decoder, and a middle block that acts as a bottleneck. The blocks of the encoder are composed of two convolutional layers followed by a Max-Pooling and a Batch Normalization layer. At these blocks, features are extracted from the data, and the output is downsampled to half of the input size. Instead, the blocks of the decoder are configured in an inverse way, substituting the pooling operators by upsampling operators which consecutively upscale the extracted low-resolution features into a final segmentation map of the input size. The networks were trained to minimize the dice loss in Equation \ref{eq:dice_loss} that measures the difference between the output and the ground truth. The training took 1000 iterations using Adam as the optimization algorithm with a learning rate of 1-e4.

\begin{equation}
    \label{eq:dice_loss}
    L_{dice} = -\frac{2\cdot\sum_{j}p_{j}g_{j}}{\sum_{j}p_{j}+g_{j}}
\end{equation}
where $p \in [0,1]$ is the network output, and $g \in \{0,1\}$ is the ground truth.

Unlike the experiments in the previous section, in the experiments in the supervised part, just defective samples were used for training. So the defective samples were split into train, validation and test sets following the next distribution: 300 for training and validation, and the remaining 75 for testing (4 crack images, 48 microcrack images, and 23 finger interruptions images). In addition, the defect-free samples used in the evaluation in the unsupervised part experiments were also employed to evaluate the models in this section. The final dataset for this part is shown in Table \ref{tab:dataset_supervised}.

\begin{specialtable}[H]
\caption{Dataset sample distribution for the supervised part experiment.}
\centering
\begin{tabular}{lccccc}
\toprule
\multicolumn{1}{c}{\textbf{}} &  & \textbf{Train} & \textbf{Val} & \textbf{Test} & \textbf{Total} \\ \hline
\multicolumn{2}{l}{\textbf{Defect-free}} & - & - & \textbf{375} & 375 \\ 
\multicolumn{2}{l}{\textbf{Defective}} & 232 & 68 & \textbf{75} & 375 \\
\multicolumn{2}{r}{Crack} & 14 & 4 & 4 & 18 \\ 
\multicolumn{2}{r}{Microcrack} & 152 & 50 & 48 & 240 \\ 
\multicolumn{2}{r}{Finger inter.} & 70 & 24 & 23 & 117 \\ \bottomrule
\end{tabular}

\label{tab:dataset_supervised}
\end{specialtable}

}
}

\subsubsection{Results}{

After training U-net separately with the two versions of the dataset (manual and automatic), the 75 defective and 375 defect-free samples were employed to compute the metrics and evaluate the performance of the models. The results are shown in Table \ref{tab:results_auto}. In addition to the U-net based models, the model from the previous section (i.e., f-AnoGAN-256) was also executed on the same test to validate that the supervised training with automatic labels improved the detection rate compared with the anomaly model.

\begin{specialtable}[H]
\caption{Image-level results from U-net trained on manually created labels, U-net trained on automatically created labels, and also, the results from the anomaly model used for annotation.}
\centering
\begin{tabular}{lccc}
\toprule
       \textbf{Model} &   \textbf{Recall} &\textbf{Precision}& \textbf{Specificity} \\ \midrule
    U-net w/ manual labels &  80  & 95& 99       \\
    U-net w/ auto. labels  & 93  & 81& 95    \\
    f-AnoGAN-256  & 79  & 73& 73    \\
  
    \bottomrule
\end{tabular}

\label{tab:results_auto}
\end{specialtable}

As shown in Table \ref{tab:results_auto}, both supervised models yielded higher detection rates than the anomaly detection models without incurring more False Positive cases. If supervised models are compared with each other, U-net trained with automatic labels was able to detect more defective samples (Recall of 93\%) than U-net trained on manual labels (Recall of 80\%). 

However, automatic labels made U-net have more False Positive cases, making the Precision and Specificity decrease from 95 to 81 and 99 to 95, respectively. However, using automatic labels resulted in more False Positive cases, making the Precision and Specificity values decrease from 95 to 81 and 99 to 95, respectively. Note that the increase of False Positives has a larger impact on the Precision because of the imbalance of defective and defect-free samples in the test set. 

As for the segmentation results illustrated in Figure \ref{fig:unet_second_train}, it can be seen that the defects were more thoroughly marked with U-net trained with automatic labels than with manual labels. The second and third samples in Figure \ref{fig:unet_second_train} are an example of this. However, impurities in the cells that were not taken into account during manual labeling were also detected as defects (e.g black spots under the crack in the second sample or around the finger in the third sample). This caused certain defect-free samples with such impurities to be classified as defective cells, which increased the number of False Positives resulting in an impact on the image-level metrics. Nevertheless, few defect-free samples present these False Positive cases.

\begin{figure*}[!ht]
    \centering
    
    \includegraphics[width=0.55\textwidth]{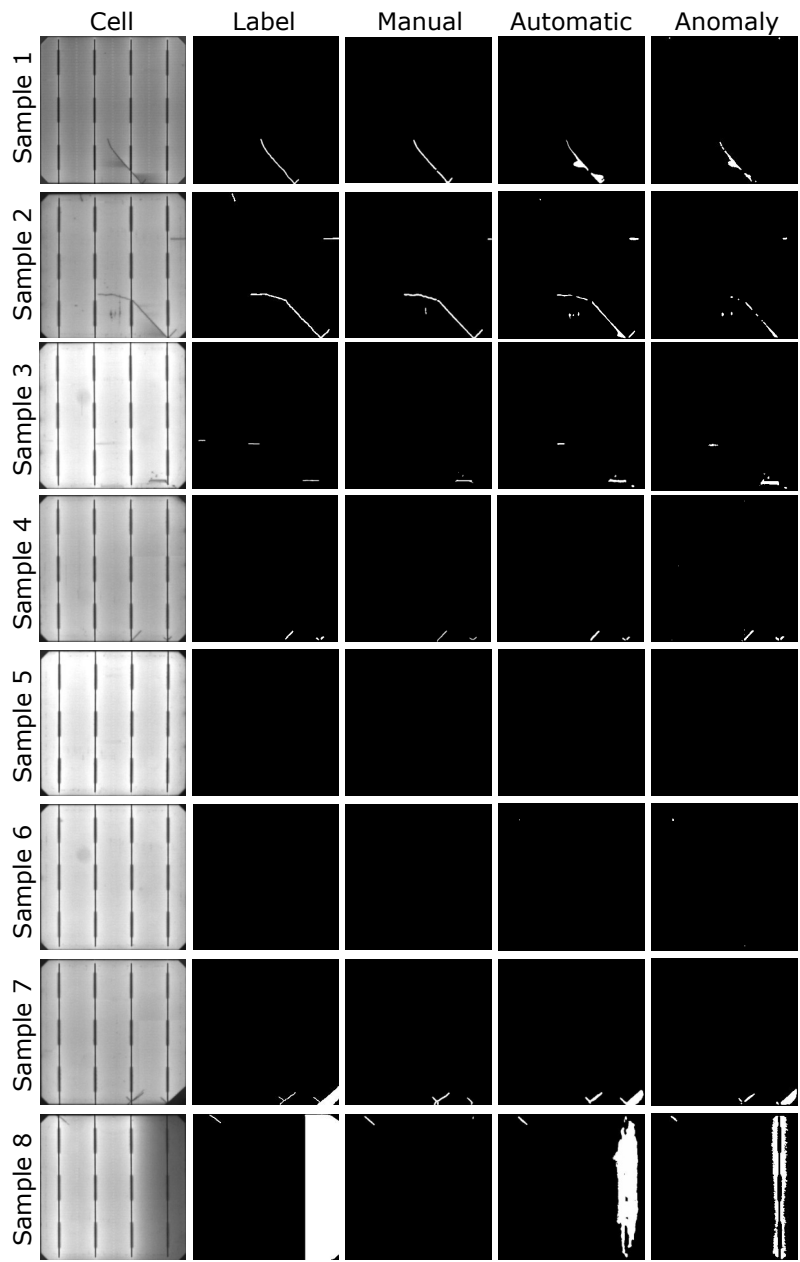}
    
    \caption{Results from supervised training models and from the anomaly model used for annotation for comparison. Samples 1, 2, 3, and 4 are defective samples with defects contemplated at training and metrics evaluation, and samples 5 and 6 are defect-free samples. Samples 7 and 8 contain defects that were not considered during training and testing, but illustrate the effect of the automatic labels in the segmentation results. Label refers to the annotation made by experts, manual refers to the segmentation results obtained from the supervised segmentation model trained with manually labeled samples, and automatic refers to the segmentation results obtained from the supervised segmentation model trained with automatically labeled samples as ground truth.}
    \label{fig:unet_second_train}
    
\end{figure*}

In addition, even not considered during training and when the metrics were calculated, the automatic labels enable U-net to segment other kinds of defects. For example, in the seventh sample in Figure \ref{fig:unet_second_train}, both models were able to detect the microcrack, but the break at the bottom right was only detected by the models trained with automatic labels. The same happened in the last row sample where the bad soldering was not segmented when using manual labels.

Concerning annotations, it seems that annotating dark areas around the defects has a positive effect on the models' pixel-level results. For example, in the first sample in Figure \ref{fig:automatic_label} the manual label does not cover the areas around the defect, whereas with automatic labeling these areas are annotated as defective. The experts did not consider these areas during the labeling as they are not part of the defect, but a consequence of the defect itself. However, these dark areas will not appear in defect-free cells. Because of that, the anomaly detection network annotated them as defective areas. When considering these dark areas as part of the labels, the network trained on automatic labels recognized dark areas around defects as defective.}

Consequently, as shown in the eighth sample in Figure \ref{fig:unet_second_train}, even if the class was not included in the training, the dark area in the right that belongs to a bad soldering defect was segmented when using automatic labels and not when using manual annotations. The same happened with the break in the first sample. Moreover, the segmentation of other defects, for example, the finger interruption in the third sample and the microcrack in the fourth, have been more accurately segmented. Nonetheless, by annotating dark areas as defective, certain impurities that were not considered as defects were also segmented. So, including dark areas as part of the labels was beneficial for pixel-level results and to detect more defective samples, even if it made some new False Positive cases arise.

\section{Conclusions}{
\label{sec:conclusions}
In this work, an anomaly detection-based methodology has been proposed for the development of a quality inspection system of monocrystalline solar cells. With anomaly detection, only defect-free samples are required to obtain a model for inspection which can detect and locate defects in the cells. This feature is key for the development of a PV module inspection system as it permits companies to have an inspection model from the very beginning stage of a new production line setup, without waiting for defective data to appear. Furthermore, it also avoids expending time in the annotation of the samples which saves a lot of effort concerning data preparation when constructing an inspection system.

In order to apply anomaly detection for industrial inspection, a GAN proposed to detect and locate anomalies in the medical domain has been adapted. The adaptations have been two-fold: First, the architecture has been modified such that the images can be processed in a single step instead of processing them by patches. In this way, less time is required to process a cell; therefore the established inspection time mark of less than half a second per cell has been met. And second, the training scheme has also been modified. This modification has resulted in an improvement in the defect detection capabilities of the model.

In addition, it has been experimentally demonstrated that the results from the anomaly detection are potential pixel-wise labels that can be used for supervised training. In the experiment, the defect localization results obtained from a model trained with labels generated by experts and a model trained with automatically generated labels have been compared. The comparison has shown that using automatic labels is comparable to using manual annotations; thus, it is feasible to use anomaly detection as an automatic annotator which saves time and resources.

The proposed methodology is rooted in the use of GANs, which are known for their difficult training process. In industry, most of the quality inspection cases are related to homogeneous parts that can alleviate to some extent the training instability that the network can face. However, less homogeneous parts might need some modifications in the training or in the network, in order to learn the data distribution and obtain high quality image reconstruction for the anomaly detection.

Lastly, although the experimental results already demonstrate the feasibility of the proposed method for inspection of solar cells, we plan to explore different architectures and parameters for optimizing the methodology in future works. It would also be interesting to test it in other industrial contexts.

}

\authorcontributions{Conceptualization, J.B., L.E. and D.M. ; Methodology, J.B. and L.E.; Validation, J.B.; Formal Analysis, J.B.; Investigation, J.B.; Data Curation, J.B.; Software, J.B.; Writing – Original Draft Preparation, J.B.; Writing – Review \& Editing, J.B., L.E. and D.M.; Visualization, J.B.; Supervision, L.E.; Funding Acquisition, L.E.}

\acknowledgments{This research was developed as part of the Elkartek project
ENSOL 2 (KK-2020/00077) supported by the Basque Government. Also, we will like to thank Jean Philippe Agerre and Jon Altube from Mondragon Assembly S. Coop. for providing us with an industrial dataset.}

\conflictsofinterest{The authors declare that they have no known competing financial interests or personal relationships that could have appeared to influence the work reported in this paper.}

\reftitle{References}


\externalbibliography{yes}

\bibliographystyle{apalike}
\bibliography{biblio_phd.bib}


\end{document}